\documentclass[10pt, conference]{IEEEtran}

\usepackage{amsmath,amssymb,amsfonts}
\usepackage{graphicx}
\usepackage{pifont}
\usepackage{setspace}
\usepackage{enumerate}

\usepackage{amsthm}
\usepackage{xcolor}
\usepackage[figuresright]{rotating}

\usepackage{graphicx}
\graphicspath{{/}{fig/}}

\usepackage{array}
\usepackage{textcomp}
\usepackage{xcolor}
\usepackage{multirow}
\usepackage{booktabs}

\usepackage{mathtools}
\usepackage{breqn}
\usepackage{float}

\usepackage{pgfplots}
\pgfplotsset{compat=1.7}
\usepgfplotslibrary{groupplots}

\usepackage[style=base]{caption}
\usepackage{subcaption}

\usepackage{multirow,tabularx}
\usepackage{hyperref}
\usepackage{flushend}
\usepackage{algorithmic}
\usepackage[vlined, ruled, shortend]{algorithm2e}

\usepackage[capitalise]{cleveref}
\usepackage{svg}

\newbox{\bigpicturebox}

\newlength\figureheight
\newlength\figurewidth
\SetAlCapNameFnt{\footnotesize}
\SetAlCapFnt{\footnotesize}

\DeclareMathOperator*{\argmin}{argmin}

\SetKwInput{kwInit}{Initilization}

\title{
    \huge
    UWB Role Allocation with Distributed Ledger Technologies for Scalable Relative Localization in Multi-Robot Systems
}

\author{
    \IEEEauthorblockN{
        \vspace{1em}
        Paola Torrico Mor\'on\IEEEauthorrefmark{2},
        Salma Salimi\IEEEauthorrefmark{2},
        Jorge Pe\~na Queralta\IEEEauthorrefmark{2},
        Tomi Westerlund\IEEEauthorrefmark{2}
    }
    \IEEEauthorblockA{
        \normalsize
        \IEEEauthorrefmark{2}\href{https://tiers.utu.fi}{Turku Intelligent Embedded and Robotic Systems (TIERS) Lab, University of Turku, Finland}.\\
        Emails: \textsuperscript{1}\{pctomo, salmas, jopequ, tovewe\}@utu.fi\\[+6pt]
    }
}

\begin{document}

\maketitle
\thispagestyle{empty}
\pagestyle{empty}



\begin{abstract}%
    \label{sec:abstract}%
    Systems for relative localization in multi-robot systems based on ultra-wideband (UWB) ranging have recently emerged as robust solutions for GNSS-denied environments. Scalability remains one of the key challenges, particularly in ad-hoc deployments. Recent solutions include dynamic allocation of active and passive localization modes for different robots or nodes in the system. with larger-scale systems becoming more distributed, key research questions arise in the areas of security and trustability of such localization systems. This paper studies the potential integration of collaborative-decision making processes with distributed ledger technologies. Specifically, we investigate the design and implementation of a methodology for running an UWB role allocation algorithm within smart contracts in a blockchain. In previous works, we have separately studied the integration of ROS\,2 with the Hyperledger Fabric blockchain, and introduced a new algorithm for scalable UWB-based localization. In this paper, we extend these works by (i) running experiments with larger number of mobile robots switching between different spatial configurations and (ii) integrating the dynamic UWB role allocation algorithm into Fabric smart contracts for distributed decision-making in a system of multiple mobile robots. This enables us to deliver the same functionality within a secure and trustable process, with enhanced identity and data access management. Our results show the effectiveness of the UWB role allocation for continuously varying spatial formations of six autonomous mobile robots, while demonstrating a low impact on latency and computational resources of adding the blockchain layer that does not affect the localization process.

\end{abstract}

\IEEEpeerreviewmaketitle


\section{Introduction}\label{sec:introduction}

Autonomous mobile robots have been penetrating multiple industries and domains of our society. At the same time, their connectivity has been increasing and networked collaborative systems have gained importance within the field~\cite{hayat2016survey}. From construction~\cite{albeaino2021trends} and industrial warehouses~\cite{plaksina2018development} to mining, the ability for robots to operate without a user input has become more important~\cite{yang2018grand}. In order to achieve a high level of autonomy and situational awareness, localization is one of the first problems to be solved~\cite{queralta2020collaborative}. Outdoors, GNSS sensors are widely used~\cite{stempfhuber2011precise}, but multipath propagation and other inherent limitations make this solution unreliable in indoors locations~\cite{qi2020cooperative, chen2016network}. Ultra-wideband (UWB)-based localization has emerged as a robust solution for localization of mobile robots in GNSS-denied environments~\cite{shule2020uwb, xianjia2021applications}. Positioning methods based on UWB have potential also for relative localization and state estimation in multi-robot systems~\cite{xu2020decentralized, xu2021omni, queralta2020vio}. However, most systems rely on active UWB ranging for localization, which presents inherent scalability limitations owing to the required time allocation of the ranging messages. In a recent work, we have presented an approach to scalable localization in mobile multi-robot systems through dynamic role allocation, combining both active and passive positioning~\cite{moron2022towards}. In this paper, our objective is to design and implement a methodology for integrating the UWB role allocation algorithm as a secure and distributed collaborative decision-making problem with blokckchain technology and smart contracts.

\begin{figure}
    \centering
    \includegraphics[width=.48\textwidth]{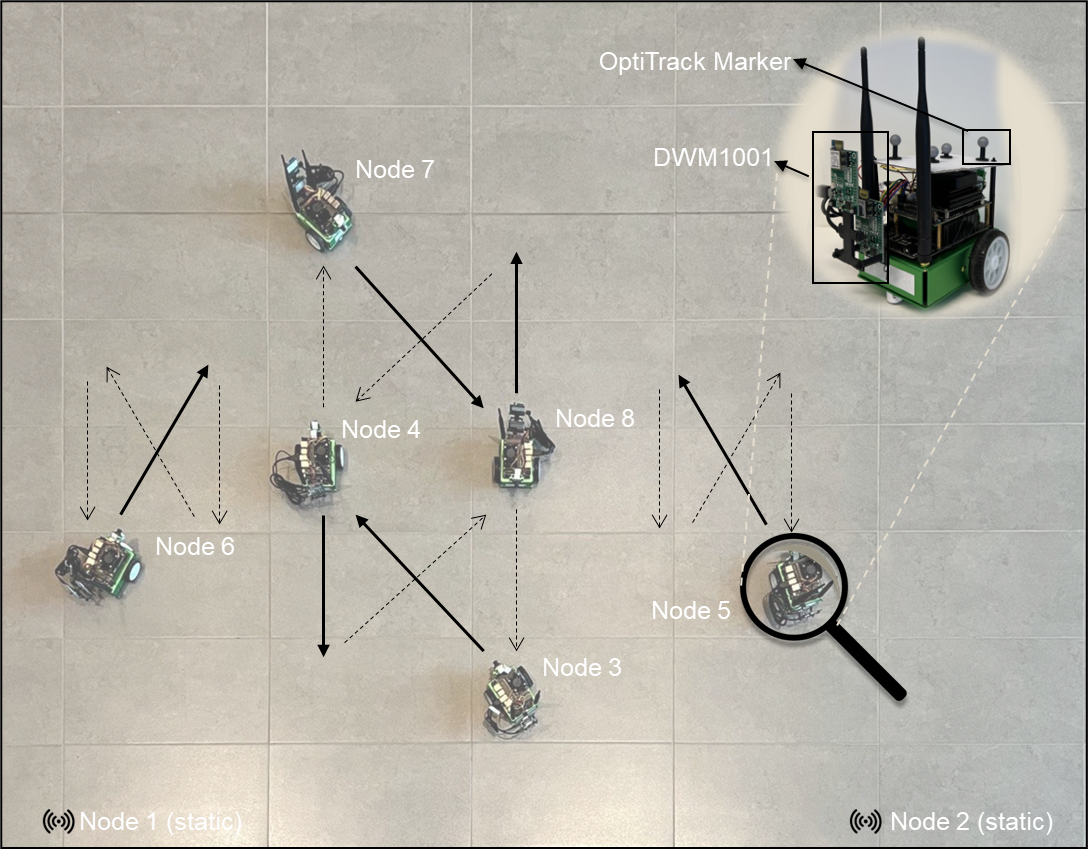}
    \caption{Experiment setting: we evaluate a dynamic role allocation approach for scalable UWB localization with smart contracts.}
    \label{fig:concept}
    \vspace{-1em}
\end{figure}

The increasing research interest towards robot swarms and multi-robot systems has brought an increasing awareness in areas such as security of robotic systems~\cite{kirschgens2018robot, mayoral2022robot}, or the explainability and auditability of their behaviour~\cite{queralta2020enhancing}. At the same time, distributed ledger technologies (DLTs) are penetrating the design and development of networked systems and the next-generation internet. At the intersection of these two trends, there is significant potential for the design of more secure and trustable distributed robotic systems, from swarms of robots~\cite{ferrer2021following} to autonomous vehicles~\cite{jain2021blockchain}. Surveillance or industrial application areas are an area of particular potential~\cite{de2021towards}.

In summary, this manuscript delves into the integration of the Hyperledger Fabric Blockchain and smart contracts with distributed role allocation methods for cooperative UWB-based localization in multi-robot systems. The main contributions of this work are:
\begin{enumerate}[i.]
    \item the design and implementation of smart contracts for managing dynamic UWB localization roles in multi-robot systems, and
    \item the deployment and experimentation of the proposed solution, extending the experimental setup of the previous work where the scalable UWB role allocation algorithm is introduced to a larger number of mobile robots~\cite{moron2022towards}.
\end{enumerate}

The remainder of the paper is structured as follows. Section II introduces related works in scalable UWB positioning in addition to spatial coordination approaches incorporating UWB-based localization and DLTs for decision-making. Section III then presents the system architecture, with the implementation details given in Section IV. Section V reports experimental results while Section VI concludes the work and lists future research directions.


\section{Background and Related Works} \label{sec:related_work}

Localization in mobile robotics utilizing radio-based technology is not a new approach. A recent trend, however, is the adoption of UWB-based solutions, specially for GNSS-denied environments~\cite{shule2020uwb}, driven by unprecedented accuracy in the field of radio-based indoor positioning at low cost. UWB provides a higher accuracy, along with lower interference in the bands it operates and better robustness against multipath when comparing with Wi-Fi or Bluetooth, allowing for a wider adoption~\cite{chen2016network}.

Autonomous mobile robots are part of multiple industries such as warehouse management ~\cite{li2016tracking}, mining~\cite{losch2018design}, packet deliveries. All these applications might not be able to provide controlled environments for the robots to operate in. Approaches based on light based sensors like cameras, LiDARS and VIO cameras depend on the quality of the environment, being susceptible to dust and low level visibility conditions ~\cite{xianjia2021applications}. Wireless technologies such as UWB are not influenced by those conditions. In addition, onboard odometry approaches using LiDar and VIO cameras might drift over time, and integrating them with UWB measurements aid it reducing and correcting long-term drift~\cite{xianjia2021applications, xianjia2021cooperative}.

In the case of UAVs, the centimeter-level accuracy that can be provided by UWB makes a cost effective replacement for a MOCAP system~\cite{guo2016ultrawideband, queralta2020uwb}, and has also the advantage that it does not need to be tied to such as a constraint environment as MOCAP systems usually are. Typical applications for UWB already in use for UAVs include ground to air ranging ~\cite{khawaja2019uwb} and  level of swarm-wide state estimation with a UWB in each robot~\cite{xu2020decentralized}.

\subsection{UWB localization}

UWB localization can be performed utilizing different ranging modalities, which are common to wireless ranging systems~\cite{xianjia2021applications}. Two widely use rangings are Time of Flight (ToF) and Time Difference of Arrival (TDoA) also known as hyperbolic positioning~\cite{hamer2018selfcalibrating}. Both are widely used in both research and industry. Angle of arrival (AoA), can also be used though it is more rare~\cite{gao2009particle}. Received Signal Strength Indicator (RSSI) is commonly used in Bluetooth technology~\cite{chen2016network, gao2009particle}.However, UWB localization systems are able to produce, out-of-the-box, centimeter-level accuracy while RSSI is more dependent on the environment and is typically at least an order of magnitude less accurate. 

Time of flight ranging measures the amount of time it takes for the signal to propagate from the initiator, that start the transmission, to the receiver~\cite{qi2020cooperative}. ToF measurements are often performed one to one between two devices, in a process called Two Way Ranging (TWR). It can either consist on two transmission, or single sided two-way ranging (SS-TWR), or three, with a double sided two-way ranging (DS-TWR)~\cite{xianjia2021applications}. In SS-TWR, a device A initiates the transmission, having device B receive it. Then B responds to the polling message, returning the message after an internal delay for processing. When device A receives the response message, it is able to compute the time of flight of the signal. On the other hand, with DS-TWR, the devices take turns initiating the communication, making it equivalent to two SD-TWR ranging~\cite{shule2020uwb}. Given the time of flight of the signal and knowing the speed of the signal in the transmission medium, the distance between devices A and B is obtained. To obtain the position of a UWB node using ToF ranging, the distance to at least three nodes is needed~\cite{guo2017ultrawideband}: once the distances are obtained, with anchors normally fixed in place, the actual position of the node can be calculated using trilateration or multilateration.

Time Difference of Arrival measures the difference of propagation time between the transmission point of a UWB signal and two or more receivers~\cite{shule2020uwb}. Then it is possible to obtain the difference of distance between the transmitter and each of the receivers. In passive TDoA scheme, a number of UWB nodes engage in constant communication and a passive UWB just receives all the messages, enabling the computation of all the differences of distances to pairs of active nodes. From these, the position can be calculated at the intersection of a series of hyperbolas. 

When deploying a system in larger areas for a larger number of nodes, the scalability of the system can be considered as a key parameter~\cite{xianjia2021applications}. The number of nodes which are actively communicating will limit the communication frequency~\cite{corbalan2018concurrent}. With ToF, the more nodes on the system the lower the frequency~\cite{heydariaan2020anguloc}. Passive TDoA does not present this problem, since part of the nodes just have to receive and not transmit any message~\cite{vecchia2019talla}. However, the main drawback is the requirement  of clock synchronization~\cite{grosswindhager2019snaploc} and the positioning error increases when leaving the convex envelope defined by the active nodes ~\cite{jansch2020robust}.

\subsection{Hyperledger Fabric}

In distributed multi-robot systems, the implementation of blockchain technology can bring some key properties, such as a built-in security while data sharing in the network~\cite{sankar2017survey}, enabling auditability through the immutability of the data~\cite{white2019black}, and collaborative decision-making through smart contracts enabled by consensus protocols ~\cite{wang2019survey, nguyen2019blockchain}. Blockchain platforms handle user credentials differently, creating with both permissionless and permissioned solutions~\cite{queralta2021blockchain}. 

Hyperledger Fabric is a private permissioned blockchain developed in Go, Java and Javascript~\cite{salimi2022towards}. It is able to achieve data isolation by enabling different channels of communication between subsets of peers, allowing for private communication~\cite{queralta2020enhancing}, ~\cite{aditya2021survey} .Hyperledger fabric provides high modularity, configurability and pluggable features, that enable developers to implement different technologies, such as consensus protocols~\cite{alvares2021blockchain}.

Having being designed for enterprise use since the beginning, Hyperledger Fabric has a set of characteristics that can be implemented in distributed robotic systems. All the participants in the blockchain are identified, which enables identity and data access management. Fabric contains tools for certificate generation and identity management. The network is permissioned, adding a layer of security to the system. It is also capable of meeting the needs of real time robotic system due to its high transaction throughput performance. Transaction confirmation can be configured so consensus can happen with low-latency but in a non deterministic manner, depending on connectivity~\cite{salimi2022towards}.


\section{System Overview}

\begin{figure}
    \centering
    \includegraphics[width=.45\textwidth]{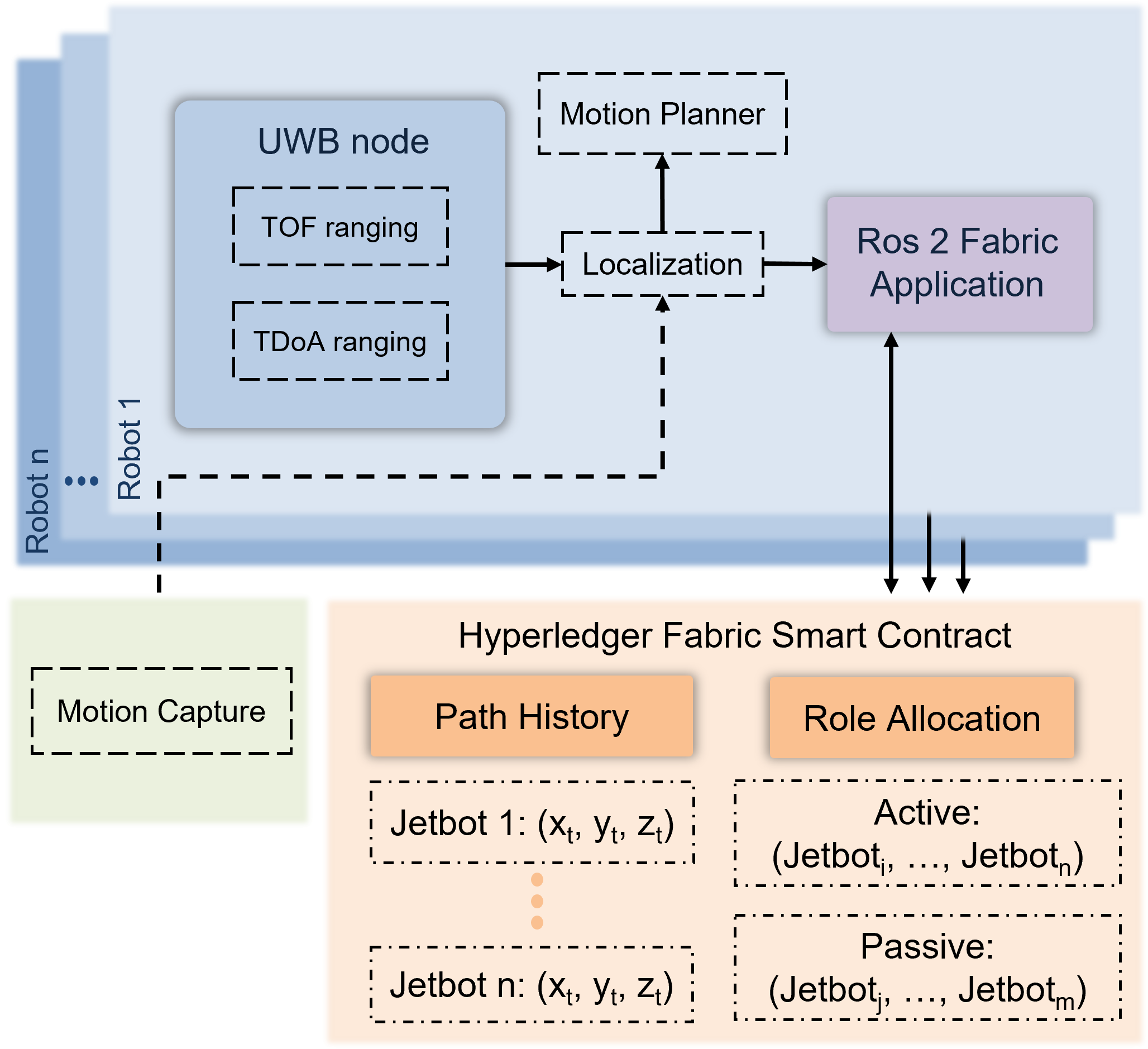}
    \caption{System overview. Each robot is able to localize itself relative to the other robots using either active ToF ranging or passive TDoA ranging through a UWB node. The motion planner follows a predefined trajectory. The localization module is interfaced with a Hyperledger Fabric ROS\,2 Go application that implements the UWB role allocation through smart contracts. A motion capture system is used to validate the experimental results.}
    \label{fig:framework_architecture}
\end{figure}

Through this section we describe the system overview. It is divided in two main sections, the UWB part for the relative localization of the robots on the system and the implementation of smart contract in the Hyperledger Fabric for the positions and roles in the system.

\subsection{UWB Relative Localization}

As mentioned before, UWB technology can be used for localization with centimeter level accuracy. It has also several ranging methods, two of the most used being ToF and TDoA. ToF localization is based on the exchange of messages between a pair of nodes, where the distance between the nodes can be measured based on the time of flight of the message. With a minimum of three distances (towards anchors), multilateration can be performed and relative localization with reference to the anchors can be obtained. The disadvantage here is that the position of some of the nodes has to be known, and more importantly, there are a high number of messages needed to be exchanged. With each new node in the system decreasing the frequency of ranging for all the UWB nodes. 

In case of TDoA, and more specifically in passive TDoA, less messages can be exchanged. Normally fixed nodes transmit ranging between them, in our particular case message for TWR, and passive nodes are able to listen to those. This time instead of computing the distance, it is the difference of distance to a pair of nodes, so TDoA is a more scalable solution than ToF, because adding more nodes does not necessarily decrease the localization frequency of the whole system. The disadvantage however, is that the localization accuracy decreases outside the convex envelope of the fixed anchors. If we consider that most or all of the nodes of the system are mobile, the probability of some passive node leaving the convex envelope is very high.

With the aim of using the strength of both ToF and TDoA, and minimize their disadvantages, we proposed a combined solution in a previous work~\cite{moron2022towards}. A subset of nodes, active nodes, use ToF for constant two way ranging and localization. A second subset of nodes, in contrast, use TDoA, receiving all the messages from the active nodes and using them for their own localization. These nodes are thus called listener nodes. In addition, the listener nodes need to be placed towards the center of the convex envelope created by the active nodes to improve accuracy. Assuming all nodes are mobile, the role of the nodes will therefore change based on the position of the nodes on the system.

\subsection{Role allocation}

The algorithm for role allocation that we integrate within a smart contract in Hyperledger Fabric is shown in Algorithm~\ref{alg:localization_algorithm} (please see our previous work~\cite{moron2022towards} for reference on the notation). The core idea behind this role allocation process is to keep listener nodes towards the centroid of the system's convex envelope, and active nodes towards the outside of the system.The implemented smart contract analyzes all the possible combinations for a fixed number of nodes and a given number of active nodes. The number of active nodes is directly derived from the desired localization frequency, based on the ranging speed (maximum ranging frequency between two nodes $f_{UWB,max}$) and the minimum localization frequency ($f_{LOC,min}$): $k=floor(f_{UWB,max}/f_{LOC,min})$. From this, the centroid of all node combinations is computed and a minimization function is used to look up which combination of nodes creates the least amount of distance error squared of the remaining nodes. The combination of nodes for the minimum are then the active nodes and the rest the passive nodes.

\begin{algorithm}[t]
    \footnotesize
	\caption{UWB role allocation and relative positioning}
	\label{alg:localization_algorithm}
	\KwIn{\\
	    \hspace{1em}Maximum number of active nodes: $k$ \\
	    \hspace{1em}Node positions at $t-1$: $\{p_i[t-1]\}\in\mathbb{R}^{3n}$; \\
	}
	\KwOut{\\
	    \hspace{1em} Set of active nodes : $A[t+1] \in \mathcal{P}_k([n])$; \\
	    \hspace{1em} Set of passive listeners : $L[t+1] = [n] \setminus A[t+1]$; \\
	    \hspace{1em} Active node positions : $\{p_{a}[t]\}_{a\in A}\in\mathbb{R}^{3k}$; \\
	    \hspace{1em} Passive listener positions : $\{p_{l}[t]\}_{l\in L} \in\mathbb{R}^{3(n-k)}$; \\
	} 
	\SetKwFunction{Ftdoa}{tdoa\_ls\_estimator}
	\SetKwFunction{Fcost}{localization\_cost_f}
	
    \SetKwProg{Fn}{Function}{:}{}
    \Fn{\Ftdoa{$A[t]$, $\{p_i[t]\}$}}{
        \KwRet $\argmin_{p \in \mathbb{R}^3} \displaystyle\sum_{\substack{i\in A[t], \: j\in A[t] \\ i \neq j}} \left( d_{ij}^{l_i} - \left( \left\lVert p - p_{j} \right\rVert - \left\lVert p - p_{i} \right\rVert \right) \right)^2$\;
         
    }
    \Fn{\Fcost{$n$, $k$, $\{p_i[t]\}$}}{
        \KwRet $\displaystyle\sum_{\substack{i\in A, \: j\in A \\ i \neq j}} \left( d_{ij}^{l_i} - \left( \left\lVert p - p_{j} \right\rVert - \left\lVert p - p_{i} \right\rVert \right) \right)^2$\;
         
    }
	\BlankLine
    \If {t = 0} {
        $\{d_{ij}\}_{i,j\in[n],i\neq j}$ $\leftarrow$ get\_tof\_ranges$\left([n]\right)$;\\
        $\{p_i[0]\}_{i\in[n]}$ $\leftarrow$ multilateration$\left(\{d_{ij}\}\right)$;\\
    }
    \Else {
        $\{d_{ij}\}_{i,j\in A,i\neq j}$ $\leftarrow$ $get\_tof\_ranges\left(A[t]\right)$;\\
        $\{d^{l}_{ij}\}_{l\in L,i,j\in A,i\neq j}$ $\leftarrow$ $get\_tdoa\_ranges\left(L[t]\right)$;\\[+0.42em]
        \tcp{Active node positions}
        $\{p_A[t]\}_{a\in A} \:\:\leftarrow \:\: multilateration\left(\{d_{ij}\}\right)$;\\[+0.42em]
        \tcp{Passive listener positions}
        $\{p_l[t]\}_{l\in L}$ $\leftarrow$ $tdoa\_ls\_estimator\left(\{d^{l}_{ij}\}\right)$;\\
    }
    \BlankLine
    \tcp{Role allocation}
    $A[t+1] \leftarrow \argmin_{A \in \mathcal{P}_k\left([n]\right)} localization\_cost\_f\left(n,k,\{p_i[t]\}\right)$;\\
    $L[t+1] \leftarrow [n] \setminus A[t+1]$

\end{algorithm}
\subsection{Hyperledger Fabric}

The aim of using Hyperledger Fabric in this paper is for secure data management and robots control. On the other hand, smart contracts in Fabric blockchain have been used for storing path history and also role allocation of the robots.

Two smart contracts have been implemented in this framework: one for storing the path tracking of six Jetbots and one for implementing the role allocation algorithm and also storing the roles. The calculated roles are then published for the UWB nodes to switch roles accordingly.


\section{Experimental Setup}

For the experiment, we used six mobile robots and two additional fixed nodes. Each of the robots moved with a predefined trajectory and a series of waypoints to reach. The motion planning is implemented in a way that robots only navigate towards a waypoint when all of them have reached their current objective. The different waypoints were selected to enforce that the robots would need to switch between inner positions and outer ones, effectively requiring continuous reallocation of roles. This allows us to analyze different metrics on the role allocation process. All six robots are equipped with a DWM1001-Dev UWB node, while two extra static UWB nodes define the frame of reference for the relative localization. The path history and roles of the nodes is stored in  Hyperledger fabric, to analyze the overhead that it adds to the system, and thus study the potential of this approach for enhanced security and trustability. The robots are deployed in an area of approximately $20\,m^2$.

\subsection{Hardware}

The platforms used in the experiment are six Jetbot robots from Waveshare. Each Jetbot incorporates a Jeson Nano Module with 4GB of RAM, 64-bit quad-core ARM A57 processor, and a Dual Band Wireless-AC 8265 for Wifi and Bluetooth connectivity. We program the Decawave's DWM1001 UWB modules with a custom firmware to enable both ToF and TDoA localization, as well as scheduling of the ToF transmissions in time. The modules are able to act as active or passive nodes. Active nodes broadcast their own ranging or position estimations, while passive nodes do so though ROS topics. Therefore, all information is available to all nodes. While this requires slightly more communication, the overhead is mostly negligible and it allows for the position of all nodes to be known through the multi-robot system.

For ground reference a MOCAP system is used, consisting of six Optitrack cameras connected to the backbone Wifi network. We use as a controller for the entire system a computer with an Intel i7-10750H processor, to coordinate the actions of the six Jetbots, run the different location algorithms, as well as participate in the Hyperledger Fabric blockchain. The robots do not directly run the role allocation smart contract but instead only an interfacing ROS\,2-Fabric application.

\subsection{Software}

The Jetbots run ROS Melodic under Ubuntu 18.04. The computer that serves as controller coordinates the waypoints for the Jetbots, publishing the next waypoint for each robot in ROS topics. The path controller, run directly in the Jetbot, subscribes to the topic and moves the it to it, waiting for the next one. The computer waits for all nodes to reach their next waypoint and then publishes a new set of objectives. The Decawave DWM1001 modules are connected to each Jetbot through a serial interface, and run custom firmware developed in C.
Each Jetbot runs a python interface to publish the information provided by the UWB to different topics.
The main controller can then either compute the position of all the nodes and also the roles directly, or use the ROS\,1-ROS\,2 bridge
to pass the necessary information to the Fabric smart contract and other ROS\,2 nodes.

We utilize the \textit{ros1\_bridge} to forward bidirectional data between the ROS\,1 nodes that interface with the Jetbot motion planner, UWB nodes and global controller, and the ROS\,2 nodes in Go that interface with the Fabric application. To simplify the setup, all topics are made available to both ROS\,1 and ROS\,2 nodes. In order to analyze the impact of the Fabric integration, we implement the role allocation algorithm in three different process: (i) a ROS\,1 Python node; (ii) a ROS\,2 Go node using \textit{rclgo}; and (iii) a Fabric smart contract also using Go. All three implementations run at a fixed rate of 5\,Hz. The role allocation algorithm does not need to run at high frequencies as it only affects the higher-frequency motion planning of the robots indirectly. The ROS\,2-Fabric application subscribes to all position data, which is stored in the blockchain through a smart contract. The role allocation results from a separate smart contract are also saved in the blockchain.

From the perspective of the Fabric blockchain setup, we deploy a private Hyperledger Fabric network. The network setup process consists of the following steps: \textbf{(1) Identities:} we define six organizations, one for each of the mobile robots (representing potentially different but collaborating parties or owners), and one orderer node. Each organization is defined with a single peer and a certificate authority (CA) for simplicity. \textbf{(2) Deployment:} Docker containers for each of the peer nodes and the ordered are deployed across the network, and we initialize the genesis block. \textbf{(3) Channels:} for the purpose of demonstrating a proof-of-concept in integrating Fabric smart contracts for collaborative decision making, we create a single channel. With the channel genesis block created, all peer nodes join the same channel. \textbf{(4) Smart contracts:} we deploy the two different chaincodes for data recording and role allocation (smart contracts) through one of the peers. The rest of the peer nodes then proceed to approve the chaincodes for their respective organizations. \textbf{(5) Invocation:} once the chaincode is approved by enough members in the channel, it can be invoked by any of the peers for data storage. A ROS\,2-Fabric Go application invokes the role allocation chaincode at the predefined 5\,Hz frequency. The application provides to the chaincode the latest known location to all nodes, received from the different ROS\,2 topics.


\begin{figure*}
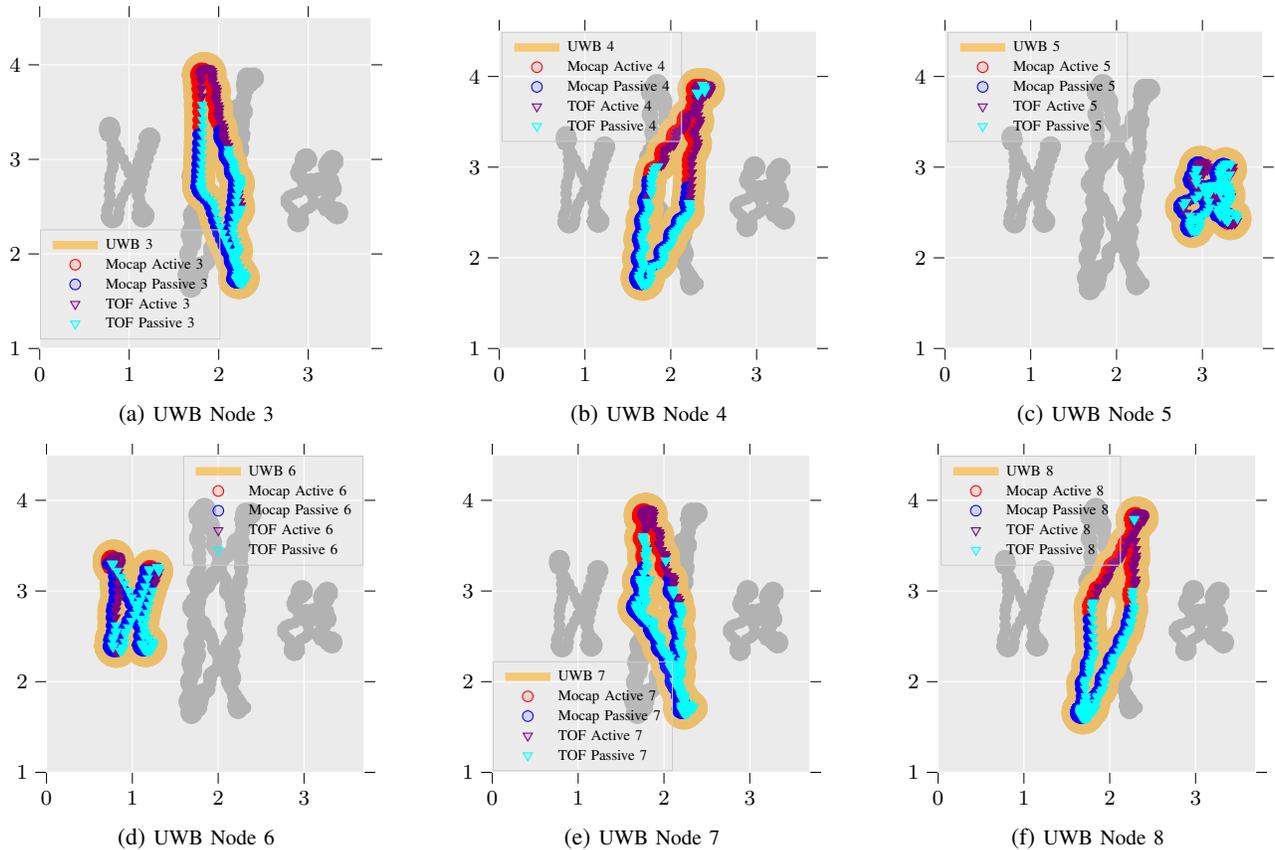


    \centering
    \begin{subfigure}[t]{0.33\textwidth}
        \centering
        \setlength\figureheight{\textwidth}
        \setlength\figurewidth{\textwidth}
        \footnotesize{\input{tex/nodes_pose3_3}}
        \caption{UWB Node 3}
    \end{subfigure}
    \begin{subfigure}[t]{0.32\textwidth}
        \centering
        \setlength\figureheight{\textwidth}
        \setlength\figurewidth{\textwidth}
        \footnotesize{\input{tex/nodes_pose3_4}}
        \caption{UWB Node 4}
    \end{subfigure}
    \begin{subfigure}[t]{0.32\textwidth}
        \centering
        \setlength\figureheight{\textwidth}
        \setlength\figurewidth{\textwidth}
        \footnotesize{\input{tex/nodes_pose3_5}}
        \caption{UWB Node 5}
    \end{subfigure}
    
    $ $ \\[+6pt]
    
    \begin{subfigure}[t]{0.32\textwidth}
        \centering
        \setlength\figureheight{\textwidth}
        \setlength\figurewidth{\textwidth}
        \footnotesize{\input{tex/nodes_pose3_6}}
        \caption{UWB Node 6}
    \end{subfigure}
    \begin{subfigure}[t]{0.32\textwidth}
        \centering
        \setlength\figureheight{\textwidth}
        \setlength\figurewidth{\textwidth}
        \footnotesize{\input{tex/nodes_pose3_7}}
        \caption{UWB Node 7}
    \end{subfigure}
    \begin{subfigure}[t]{0.32\textwidth}
        \centering
        \setlength\figureheight{\textwidth}
        \setlength\figurewidth{\textwidth}
        \footnotesize{\input{tex/nodes_pose3_8}}
        \caption{UWB Node 8}
    \end{subfigure}
    
    \caption{Plots with the trajectories of each of the six moving Jetbots. The orange paths show the predefined trajectory of the Jetbot. The red and blue marks show the roles that the Jetbot would be allocated to, if we were using the positioning data from the MOCAP system as a reference. The red color represents an active role while blue represents a passive role. The purple and cyan marks show, respectively, the same active and passive roles calculated using ToF information from the UWB transceivers and without the external reference.}
    \label{fig:experiment_results}
    
\end{figure*}

\begin{table}
    \centering
    \caption{The table shows the percentage of overlap in the roles when using the MOCAP localization vs the ToF ranging localization}
    \small
    \begin{tabular}{@{}lccccc@{}}
        \toprule
        & \multicolumn{2}{l}{\textbf{\quad MOCAP[\%]}}     & \multicolumn{2}{l}{\textbf{\quad \  ToF[\%]}}   &\multirow{2}{*}{\textbf{\begin{tabular}[c]{@{}l@{}}Total \% \\ Overlap\end{tabular}}} \\
        & \textbf{Active}     & \textbf{Passive} & \textbf{Active} & \textbf{Passive} & \\
        \midrule
        \textbf{Node 3}       & 25.00   & 75.00 & 27.50 & 72.50 & 94.17   \\
        \textbf{Node 4}       & 25.00   & 75.00 & 27.50 & 72.50 & 87.92   \\
        \textbf{Node 5}       & 25.00   & 75.00 & 27.50 & 72.50 & 64.17   \\
        \textbf{Node 6}       & 25.00   & 75.00 & 27.50 & 72.50 & 82.08   \\
        \textbf{Node 7}       & 25.00   & 75.00 & 27.50 & 72.50 & 93.33   \\
        \textbf{Node 8}       & 25.00   & 75.00 & 27.50 & 72.50 & 95.42   \\
        
        \bottomrule
    \end{tabular}
    \label{tab:percentage}
\end{table}

\begin{figure}
    \centering
    \setlength\figureheight{.35\textwidth}
    \setlength\figurewidth{.48\textwidth}
    \footnotesize{
\begin{tikzpicture}

\definecolor{darkgray176}{RGB}{176,176,176}
\definecolor{darkorange25512714}{RGB}{255,127,14}

\definecolor{color4}{RGB}{255,255,204} 
\definecolor{color3}{RGB}{204,255,204} 
\definecolor{color2}{RGB}{204,255,255} 
\definecolor{color1}{RGB}{204,204,255} 
\definecolor{color0}{rgb}{1,0.752941176470588,0.796078431372549} 

\begin{axis}[
width=\figurewidth,
height=\figureheight,
axis line style={white},
legend style={fill opacity=0.8, draw opacity=1, text opacity=1, 
draw=white!80!black},
tick align=outside,
tick pos=left,
x grid style={darkgray176},
xmin=0.5, xmax=3.5,
xtick style={color=black},
xtick={1,2,3,4,5},
xticklabels={Python, Go, Fabric+Go},
y grid style={darkgray176},
ymajorgrids,
ylabel={Time (ms)},
ymode=log,
ymin=6.057214419123e-05, ymax=0.077709654801712,
ytick style={color=black}
]
\addplot [black, fill=color0] 
table {%
0.75 0.00456380844116211
1.25 0.00456380844116211
1.25 0.00554710626602173
0.75 0.00554710626602173
0.75 0.00456380844116211
};
\addplot [black] 
table {%
1 0.00456380844116211
1 0.00417184829711914
};
\addplot [black] 
table {%
1 0.00554710626602173
1 0.00652503967285156
};
\addplot [black] 
table {%
0.875 0.00417184829711914
1.125 0.00417184829711914
};
\addplot 
table {%
0.875 0.00652503967285156
1.125 0.00652503967285156
};
\addplot [black, mark=o, mark size=1.23, mark options={solid,fill opacity=0}, only marks] 
table {%
1 0.00885009765625
1 0.0126910209655762
1 0.0133779048919678
1 0.00818514823913574
1 0.0080718994140625
1 0.0119750499725342
1 0.00936603546142578
1 0.00871706008911133
1 0.0122511386871338
1 0.00887393951416016
1 0.0094151496887207
1 0.0092310905456543
1 0.00915002822875977
1 0.0067589282989502
1 0.0090029239654541
1 0.0108089447021484
1 0.0181088447570801
1 0.0101470947265625
1 0.00791287422180176
1 0.00825405120849609
1 0.00801897048950195
1 0.00921893119812012
1 0.0117990970611572
1 0.0115029811859131
1 0.00918698310852051
1 0.00734710693359375
1 0.0162289142608643
1 0.0161049365997314
1 0.0229790210723877
1 0.00798606872558594
1 0.009674072265625
1 0.00746583938598633
1 0.0115160942077637
1 0.00709390640258789
1 0.0066840648651123
1 0.00919079780578613
1 0.0108201503753662
1 0.0131909847259521
1 0.00861406326293945
1 0.00914192199707031
1 0.00685405731201172
1 0.00865697860717773
1 0.00762796401977539
1 0.00660896301269531
1 0.0143721103668213
1 0.0141861438751221
1 0.0206050872802734
1 0.00924992561340332
1 0.00686502456665039
1 0.0127220153808594
1 0.00708794593811035
1 0.0170769691467285
1 0.0107800960540771
1 0.0120201110839844
1 0.0095369815826416
1 0.00775790214538574
1 0.00953888893127441
1 0.00854396820068359
1 0.00755691528320312
1 0.0072319507598877
1 0.00892090797424316
1 0.0119240283966064
1 0.00838804244995117
1 0.00912094116210938
1 0.0072779655456543
1 0.00839710235595703
1 0.00689697265625
1 0.00696110725402832
1 0.00802493095397949
1 0.00961208343505859
1 0.00883293151855469
1 0.0144319534301758
1 0.00951790809631348
1 0.00858616828918457
1 0.0071101188659668
1 0.0093238353729248
1 0.0317990779876709
1 0.0106298923492432
1 0.00937199592590332
1 0.00890398025512695
1 0.0104579925537109
1 0.00765895843505859
1 0.0126838684082031
1 0.0077359676361084
1 0.00822210311889648
1 0.00911808013916016
1 0.00915193557739258
1 0.0090639591217041
1 0.00811314582824707
1 0.0136041641235352
1 0.00807499885559082
1 0.00680899620056152
1 0.0143880844116211
1 0.0155529975891113
1 0.0115950107574463
1 0.0078129768371582
1 0.00889492034912109
1 0.00654816627502441
1 0.00737404823303223
1 0.00808310508728027
1 0.00886917114257812
1 0.0193600654602051
1 0.00674986839294434
1 0.00789618492126465
1 0.0113670825958252
1 0.00745511054992676
1 0.0183959007263184
1 0.0185339450836182
1 0.00699520111083984
1 0.0119349956512451
1 0.0171740055084229
1 0.0077059268951416
1 0.00746703147888184
1 0.0137689113616943
1 0.00842618942260742
1 0.00999307632446289
1 0.0118179321289062
1 0.00769710540771484
1 0.00857806205749512
1 0.00659918785095215
1 0.00728392601013184
1 0.0112857818603516
1 0.0161011219024658
1 0.00659799575805664
1 0.0155370235443115
1 0.00804495811462402
1 0.0101110935211182
1 0.00896906852722168
1 0.0134251117706299
1 0.0109012126922607
1 0.00660204887390137
1 0.0145530700683594
1 0.00678086280822754
1 0.00859212875366211
1 0.00731301307678223
1 0.00711989402770996
1 0.00669598579406738
1 0.0125410556793213
1 0.00857400894165039
1 0.00670909881591797
1 0.00954604148864746
1 0.00664997100830078
1 0.0118339061737061
1 0.00688409805297852
1 0.0135140419006348
1 0.00729703903198242
1 0.0157878398895264
1 0.0106968879699707
1 0.0159838199615479
1 0.0207819938659668
1 0.0105500221252441
1 0.00705695152282715
1 0.0174510478973389
1 0.00653910636901855
1 0.00921511650085449
1 0.0170509815216064
1 0.0091710090637207
1 0.00666713714599609
1 0.00851702690124512
1 0.00923800468444824
1 0.0102889537811279
1 0.00734710693359375
1 0.010221004486084
1 0.00715899467468262
1 0.0206048488616943
1 0.0121510028839111
1 0.00709295272827148
1 0.0126380920410156
1 0.0124199390411377
1 0.0102801322937012
1 0.00824403762817383
1 0.00904083251953125
1 0.00929903984069824
1 0.00656700134277344
1 0.00666999816894531
1 0.0108439922332764
1 0.00765109062194824
1 0.00862002372741699
1 0.0078580379486084
1 0.0084528923034668
1 0.00846385955810547
1 0.0119419097900391
1 0.00938796997070312
1 0.0107429027557373
1 0.0140471458435059
1 0.00745105743408203
1 0.0100710391998291
1 0.00928187370300293
1 0.00819611549377441
1 0.011146068572998
1 0.00927090644836426
1 0.00835323333740234
1 0.00747799873352051
1 0.00710010528564453
1 0.00752520561218262
1 0.0110838413238525
1 0.011970043182373
1 0.00835394859313965
1 0.011059045791626
1 0.00722599029541016
};
\addplot [black, fill=color1] 
table {%
1.75 0.000107371
2.25 0.000107371
2.25 0.000133522
1.75 0.000133522
1.75 0.000107371
};
\addplot [black]
table {%
2 0.000107371
2 8.386e-05
};
\addplot [black]
table {%
2 0.000133522
2 0.000159533
};
\addplot [black]
table {%
1.875 8.386e-05
2.125 8.386e-05
};
\addplot [black]
table {%
1.875 0.000159533
2.125 0.000159533
};
\addplot [black, mark=o, mark size=1.23, mark options={solid,fill opacity=0}, only marks]
table {%
2 0.000189896
2 0.000205365
2 0.000169599
2 0.00018575
2 0.000180623
2 0.000165571
2 0.000186522
2 0.000172317
2 0.000184629
2 0.000163078
2 0.000191823
2 0.000162096
2 0.000165357
2 0.000176431
2 0.000204523
2 0.000165279
2 0.000163911
2 0.000162379
2 0.00018851
2 0.000163519
2 0.000345949
2 0.00078709
2 0.000159701
2 0.000170869
2 0.000162182
2 0.000165171
2 0.000173583
2 0.000181727
2 0.000198304
2 0.00016556
2 0.000168664
2 0.000160158
2 0.000174187
2 0.000201583
2 0.000198408
2 0.000175264
2 0.000165783
2 0.000223889
2 0.000169498
2 0.000542138
2 0.000160152
2 0.000187236
2 0.000162481
2 0.000169002
2 0.000161522
2 0.000176434
2 0.000159772
2 0.000162268
2 0.000169045
2 0.000173667
2 0.00016078
2 0.000169507
2 0.000167124
2 0.000366318
2 0.0001752
};
\addplot [black, fill=color2]
table {%
2.75 0.031153522
3.25 0.031153522
3.25 0.041125003
2.75 0.041125003
2.75 0.031153522
};
\addplot [black]
table {%
3 0.031153522
3 0.029979439
};
\addplot [black]
table {%
3 0.041125003
3 0.050760623
};
\addplot [black]
table {%
2.875 0.029979439
3.125 0.029979439
};
\addplot [black]
table {%
2.875 0.029979439
3.125 0.029979439
};
\addplot [black, mark=o, mark size=1.23, mark options={solid,fill opacity=0}, only marks]
table {%
3 0.053951208
3 0.056129745
3 0.053088298
3 0.05561578
3 0.055491688
3 0.05227508
};
\addplot [thick, black]
table {%
0.75 0.00492608547210693
1.25 0.00492608547210693
};
\addplot [thick, black]
table {%
1.75 0.000118488
2.25 0.000118488
};
\addplot [thick, black]
table {%
2.75 0.032302987
3.25 0.032302987
};

\end{axis}

\end{tikzpicture}}
    \caption{The latency of the role allocation algorithm for Python, Go and Fabric+Go implementation when running the algorithm at 5Hz.}
    \vspace{-1em}
    \label{fig:role_latency}
\end{figure}
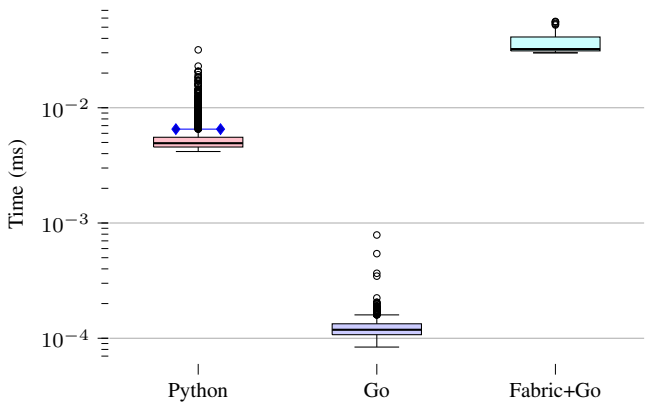

\section{Experimental Results}

This section discusses the results obtained through our experiments. The results report the role allocation results based on data calculated with the UWB nodes mounted on the six mobile Jetbots, as well as the reference MOCAP system. We also report the latency of the three different role allocation processes that we implemented (ROS\,1-Python, ROS\,2-Go and Fabric smart contract in Go) \Cref{fig:experiment_results} illustrates the different paths of the Jetbots during one of the experimental runs. Four of the mobile robots, namely nodes 3, 4, 7 and 8, share the area in which they operate and their positions continuously switch between inner and outer positions between from the perspective of the robot's convex envelope. Taking into account that the node 1 of the system is in the origin and node 2 defines the x-axis at a distance of 3.4\,m, the four nodes are passive when located towards the bottom left of the graph, the inner part of the system. Node 6 is, most of the time, an passive node, only changing to active node when located towards the upper left part of its trajectory. Node 5 is also mostly passive, tending to be in the inner part of the envelope, behind the line defined by node 2 and nodes 4 or 8.  

\Cref{fig:experiment_results} shows, importantly, a comparison between the results obtained with reference localization obtained by the MOCAP system and the localization obtained by using ToF ranging. For both localization schemes, the role allocation algorithm remains the same. The difference in the roles in both systems is due to the error from the UWB localization. However, this does not result in poorer system performance. The core takeaway is that the roles effectively change with both the reference MOCAP and the UWB localization. A system failure would occur when a passive node remains passive even when located in the convex envelope of the system and far from other inner nodes. In such case, the TDoA localization method would easily diverge and unexpected positioning information be estimated by the algorithm. \Cref{tab:percentage} shows the percentage of overlap of the MOCAP-based role allocation and the UWB-based role allocation results. We can see that five out of six robots show an overlap larger than 80\%, with three of them showing over 90\% overlap. Node 5, which has a trajectory that is inherently defined around the convex envelope, shows a lower overlap of roles with only 64.17\%. This can be explained by the uncertainty in the UWB localization being bigger than the distance between the node and the actual convex envelope in most of its trajectory. While the results are satisfactory in this case, the planning module could integrate knowledge of such a limitation in the role allocation process. Therefore, the controller can define waypoints that minimize the potential for errors in the role allocation process accounting for error estimation of the UWB localization. 

Finally, we analyze the impact of integrating Hyperledger Fabric smart contracts for the role allocation algorithm as a distributed and cooperative decision-making process by measuring the role allocation latency. The results are shown in\Cref{fig:role_latency}. The role allocation algorithm is running at 5\,Hz for all three cases. The ROS\,2 Go implementation is naturally faster than the interpreted ROS\,1 Python node. However, both implementations are fast enough for meeting the frequency defined for the role allocation. Even though the addition of a Hyperledger Fabric interface to the baseline Go implementation increments its latency, it still remains fast enough. From these results, we can conclude that it is effectively possible to run this type of distributed decision-making algorithms within the Fabric blockchain without a significant impact on the performance of the system. The benefits of Fabric for this use case include the possibility of managing identities, and the inherent immutability of data stored in the blockchain. In addition, the chaincode implementation of the role allocation algorithm enhances the trust on the results and within the system. There is potential for enabling trustable relative collaboration between multiple third parties or different robots. While different Fabric organizations are created for the experiments, we have focused on a proof-of-concept implementation where the actual control was not distributed.  Our results, nonetheless, open the door to more secure and trustable multi-robot collaboration and identity management with a permissioned blockchain, and to more scalable localization and distributed decision-making.


\section{Conclusion}\label{sec:conclusion}

In this paper we have proposed, designed and implemented a role allocation methodology for UWB-based localization relying on the Hyperledger Fabric blockchain. We analyze the performance of the role allocation algorithm with multiple moving nodes, with larger-scale experiments than in previous works. Our results show that the role allocation algorithm performs well with a higher number of mobile robots. In addition, we analyze the impact on the performance when implementing the algorithm through Fabric smart contracts, concluding that the additional latency is not significant.

In future works we seek to extend the uses of Fabric smart contracts to other collaborative decision-making problems, while extending the experiments to include aerial robots.


\section*{Acknowledgment}

This research work is supported by the R3Swarms project funded by the Secure Systems Research Center (SSRC), Technology Innovation Institute (TII).

\newpage
\bibliographystyle{unsrt}
\bibliography{bibliography}

\end{document}